\documentclass{article}

\usepackage[utf8]{inputenc}
\usepackage{scicite}
\usepackage[margin=1in]{geometry}
\usepackage[T1]{fontenc}
\usepackage{url}
\usepackage{booktabs}
\usepackage{amsfonts}
\usepackage{nicematrix}
\usepackage{graphicx}
\usepackage{pgfplots}
\usepackage{cite}
\usepackage{hyperref}
\usepackage{float}

\title{ARC Prize 2024: Technical Report}

\author{
    François Chollet,
    Mike Knoop,
    Gregory Kamradt,
    Bryan Landers
}

\begin{document}

\maketitle

\begin{abstract}
As of December 5, 2024, the ARC-AGI benchmark is five years old and remains unbeaten. We believe it is currently the most important unsolved AI benchmark in the world because it seeks to measure generalization on novel tasks – the essence of intelligence – as opposed to skill at tasks that can be prepared for in advance. This year, we launched ARC Prize, a global competition to inspire new ideas and drive open progress towards AGI by reaching a target benchmark score of 85\%. As a result, the state-of-the-art score on the ARC-AGI private evaluation set increased from 33\% to 55.5\%, propelled by several frontier AGI reasoning techniques including deep learning-guided program synthesis and test-time training. In this paper, we survey top approaches, review new open-source implementations, discuss the limitations of the ARC-AGI-1 dataset, and share key insights gained from the competition.
\end{abstract}

\section{Introduction: ARC-AGI}

Fran\c{c}ois Chollet first wrote about the limitations of deep learning in 2017~\cite{chollet2017deep}. In 2019, he formalized these observations into a new definition of artificial general intelligence (AGI), characterizing it as a system capable of efficiently acquiring new skills and solving novel problems for which it was neither explicitly designed nor trained.~\cite{chollet2019intelligence}

Alongside this definition, Chollet published the Abstraction and Reasoning Corpus (ARC) benchmark~\cite{chollet2019github} (later renamed ARC-AGI to avoid name collisions with other AI benchmarks), as a first concrete attempt to measure this definition of intelligence. We will refer to this dataset as ARC-AGI-1. It is a set of independent ``tasks'' (see figure 1), each consisting of a number of ``demonstration pairs'' (two or more, with a median count of three) and one or more ``test inputs''. A test pair consists of an ``input grid'', a rectangular grid of variable size (up to a maximum size of 30 rows by 30 columns) where each cell can have one of ten distinct ``values'', and an output grid which should be fully inferable from the characteristics of the input grid. The goal is to use the demonstration pairs to understand the nature of the task, and use this understanding to construct the output grid corresponding to each test input. The test taker is allowed two attempts per test input.

The defining characteristic of the benchmark is that it should not be possible to prepare for any of the tasks in advance. Every task in the dataset follows a different logic. All tasks were created by humans to ensure a high degree of novelty and diversity.

ARC-AGI tasks do not require specialized world knowledge (e.g., historical facts) nor language to solve. The only assumed prior knowledge is Core Knowledge~\cite{chollet2019intelligence} – concepts such as objectness, basic topology, elementary integer arithmetic, etc. Human Core Knowledge has been investigated by Spelke et al.~\cite{spelke2007}. These knowledge priors are acquired by children very early (typically before age four) and are universally shared by all humans. The ARC-AGI-1 public training tasks are designed to expose test-takers to all the Core Knowledge priors needed to solve ARC-AGI tasks.

\subsection{Dataset Composition}
ARC-AGI-1 consists of 1,000 tasks split into four subsets:
\begin{itemize}
    \item Public training tasks (400, easy) - Intended to demonstrate the task format and allow for learning the Core Knowledge priors.
    \item Public evaluation tasks (400, hard) - Intended to let researchers locally evaluate their performance.
    \item Semi-private evaluation tasks (100, hard) - Intended to let us evaluate third-party approaches that rely on publicly-available commercial APIs. It is ``semi-private'' because while it hasn't been publicly released, it has been exposed to commercial APIs and thus suffers from a risk of leakage.
    \item Private evaluation tasks (100, hard) - Intended to let us evaluate standalone approaches. It is fully private and theoretically free of leakage.
\end{itemize}

\begin{figure}[h]
    \centering
    \includegraphics[width=1\textwidth]{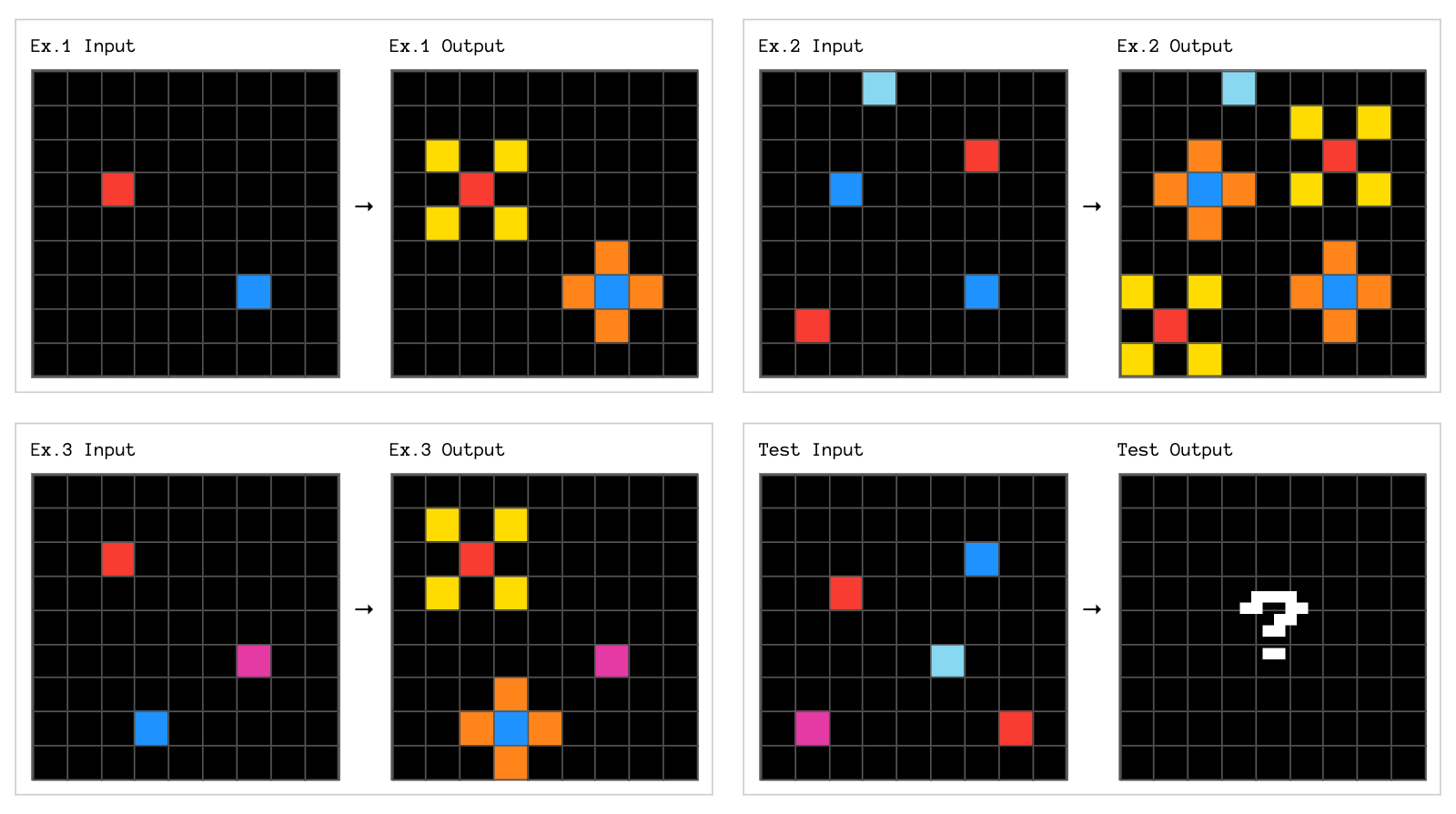}
    \caption{Example ARC-AGI task (0ca9ddb6).}
    \label{fig:example-task}
\end{figure}

State-of-the-art scores are only reported on the private evaluation task set in order to reduce the risk of overfitting and data contamination.

Another important characteristic of ARC-AGI tasks is that they are hard for AI systems, yet easy for humans. The original private evaluation tasks were originally tested by two people who scored 97\% and 98\%, and, together, solved all 100\%. NYU published a recent study testing Mechanical Turk workers which showed 99\% of public evaluation tasks were solved by at least one worker, with 10 workers assigned to each task.~\cite{nyu2024}

\vspace{0.1cm}

\subsection{Pre-2024 Progress}
ARC-AGI has been the target of three public competitions before ARC Prize 2024:
\begin{itemize}
    \item 2020: First ARC-AGI Kaggle competition (\$20,000 USD in prizes)~\cite{kaggle2020}
    \item 2022: ARCathon 1 (\$100,000 in prizes)~\cite{arcathon2022}
    \item 2023: ARCathon 2 (\$100,000 in prizes)~\cite{arcathon2023}
\end{itemize}

In the years following the original release of ARC-AGI-1, pure deep learning approaches turned out to perform poorly on ARC-AGI, as the classic deep learning paradigm works by relating new situations to situations seen at training time, with no adaptation or knowledge recombination at test time, making it impossible for such models to apprehend entirely novel tasks at test time. In the first Kaggle competition (2020), no deep-learning based approach scored above 1\%. The original GPT-3 model from OpenAI scored 0\% on the public evaluation via direct prompting. 

This is why, despite being created before large language models (LLMs), ARC-AGI has resisted the rise of LLMs in the 2022-2024 period.

The first ARC-AGI competition ran on Kaggle in 2020~\cite{kaggle2020} with a top score of 20\%. Four years later, the top score had only increased to 33\%. This lack of progress on ARC-AGI can be attributed to lack of progress towards AGI. From 2020 to early 2024, the field of AI research was dominated by the scaling up of deep learning systems, which increased task-specific skills but did not improve the ability to tackle tasks without available training data at training time (i.e., general intelligence). Our view is that progress towards AGI had stalled during this period – AI systems had been getting bigger and memorizing ever more training data, but generality in frontier AI systems had been roughly static.

\section{ARC Prize 2024 Results}

\subsection{Kaggle Leaderboard}

The poor performance of frontier AI systems on ARC-AGI at the start of 2024 was clear evidence of the conceptual limitations hindering AGI progress. In response, we launched ARC Prize~\cite{arcprize2024} to inspire AI researchers to work on new ideas and share them openly. Most frontier AI research is no longer being published by industry labs, which is why ARC Prize incentivizes and promotes open sharing.

ARC Prize 2024 launched on June 11, 2024, and ended November 10, 2024. The competition ran both on \href{https://kaggle.org}{kaggle.com} and \href{https://arcprize.org}{arcprize.org}. Prizes included a Grand Prize of \$600,000 USD for the first team to reach 85\% on the private evaluation set, \$50,000 in progress prizes tied to the Kaggle leaderboard, and \$75,000 in prizes for the best paper submissions. The Grand Prize was not claimed.

The 2024 winners are are shown in Table 1. All scores are open source and reproducible on \href{https://arcprize.org}{arcprize.org}.

\begin{table}[h]
\centering
\begin{tabular}{llr}
\toprule
\textbf{Place} & \textbf{Name} & \textbf{Score (Private evaluation set)} \\
\midrule
1st & the ARChitects & 53.5\% \\
2nd & Guillermo Barbadillo & 40\% \\
3rd & alijs & 40\% \\
4th & William Wu & 37\% \\
5th & PoohAI & 37\% \\
\bottomrule
\end{tabular}
\caption{ARC Prize 2024 Winners.}
\end{table}

The winners competed on Kaggle, where their solutions attempted to solve the 100 tasks of the private evaluation set on a virtual machine that featured a single P100 GPU in under 12 hours with no internet access. Only those who open-sourced their solution could be named as a winner and claim prizes. MindsAI achieved the highest score of 55.5\% on the private evaluation set during the competition but chose not to open source their solution, and was therefore ineligible for a prize.

\subsection{Public Leaderboard}

In addition to the Kaggle leaderboard, ARC Prize also featured a secondary leaderboard, ARC-AGI-Pub, which allowed internet access and relaxed compute constraints, in order to evaluate the performance achievable with closed-source frontier models. Due to the risk of data leakage, submissions were not verified on the private evaluation set, but rather on the ``semi-private'' evaluation set (100 tasks). We report results alongside the public evaluation set (400 tasks) to guard against overfitting. We consider scores overfit if semi-private and public evaluation set scores exceed $\pm$10\% absolute difference. All scores are open source and reproducible on \href{https://arcprize.org}{arcprize.org}.

\begin{table}[h]
  \centering
  \begin{tabular}{lcc}
    \toprule
    \textbf{Name} & \textbf{Semi-private eval} & \textbf{Public eval} \\
    \midrule
    Jeremy Berman & 53.6\% & 58.5\% \\
    Akyürek et al. & 47.5\% & 62.8\% \\
    Ryan Greenblatt & 43\% & 42\% \\
    OpenAI o1-preview (pass@1) & 18\% & 21\% \\
    Anthropic Claude 3.5 Sonnet (pass@1) & 14\% & 21\% \\
    OpenAI GPT-4o (pass@1) & 5\% & 9\% \\
    Google Gemini 1.5 (pass@1) & 4.5\% & 8\% \\
    \bottomrule
  \end{tabular}
  \caption{ARC-AGI-Pub leaderboard.}
  \label{tab:public_leaderboard}
\end{table}

This leaderboard offered entrants approximately 1,000 times more compute than the Kaggle leaderboard. ARC-AGI-Pub entries were allowed to consume up to \$10,000 in API credits, whereas entries on Kaggle could only consume the equivalent of \$10 of compute per entry. ARC Prize covered API fees for public leaderboard high score submission verification.

Final 2024 top scores on the public leaderboard are shown in Table 2. The ``pass@1'' commercial API results use a publicly available direct prompting approach (identical for all models.)

Surprisingly, both the competition and secondary public leaderboard top scores tracked closely. This suggests algorithmic improvements towards AGI hold significant power and that massive compute may not be necessary in order to beat ARC-AGI. 

\subsection{Paper Awards}

ARC Prize 2024 also featured ``Paper Awards'' to reward novel concepts regardless of how their solutions scored. Prizes were awarded to the following papers. All papers are shared alongside open source code on \href{https://arcprize.org}{arcprize.org}.

\vspace{1.2cm}

\begin{itemize}
    \item \textbf{First place}: Li et al., \textit{``Combining Induction and Transduction for Abstract Reasoning''}
    \item \textbf{Second place}: Akyürek et al., \textit{``The Surprising Effectiveness of Test-Time Training for Abstract Reasoning''} 
    \item \textbf{Third place}: Bonnet and Macfarlane, \textit{``Searching Latent Program Spaces''}
    \item \textbf{Runners up}:
    \begin{itemize}
        \item Franzen et al., (the ARChitects): \textit{``The LLM ARChitect: Solving ARC-AGI Is A Matter of Perspective''}
        \item Barbadillo, \textit{``Omni-ARC''}
        \item Fletcher-Hill, \textit{``Mini-ARC: Solving Abstraction and Reasoning Puzzles with Small Transformer Models''}
        \item Ouellette, \textit{``Towards Efficient Neurally-Guided Program Induction for ARC-AGI''}
        \item Puget, \textit{``A 2D nGPT Model For ARC Prize''}
    \end{itemize}
\end{itemize}

In total, 1,430 teams submitted 17,789 entries for ARC Prize 2024. Many well-funded startups have also shifted priorities to work on ARC-AGI -- we've heard from seven such companies this year. Additionally, multiple large corporate labs have now spun up internal efforts to tackle ARC-AGI.

While we have a long way to go to reach AGI, we're excited that ARC Prize has catalyzed several new open source frontier AGI reasoning approaches, in particular test-time training, an approach we first observed in use by Jack Cole and Mohamed Osman in 2023 and subsequently popularized this year by ARC Prize.

\section{Top Approaches}

Until 2024, all top ARC-AGI approaches relied on discrete program search, starting with the 2020 winning entry by icecuber~\cite{kaggle2020}, which exclusively leveraged brute-force program search to achieve 20\% on the private evaluation set.

Over the next four years, progress was slow. Despite the advent of LLMs (e.g., GPT-3, 3.5, 4), attempts to use these systems to beat ARC-AGI were unsuccessful. Advancement primarily came in the form of improved domain-specific languages (DSLs), notably one created by Michael Hodel~\cite{hodel2024dsl} which improved the performance of the program search process.

Progress re-accelerated, however, during ARC Prize 2024, catalyzed by three major categories of approaches:

\begin{itemize}
    \item \textbf{Deep learning-guided program synthesis:} Leveraging deep learning models, particularly specialized code LLMs, to generate task-solving programs or guide the program search process beyond blind brute-force methods.
    \item \textbf{Test-time training (TTT) for transductive models\footnote{A ``transductive'' model is a model that attempts to directly predict the output grid given a test input grid and a task specification, instead of first trying to write down a program that matches the task.}:} Fine-tuning an LLM at training time on a given ARC-AGI task specification in order to recombine the prior knowledge of the LLM into a new model adapted to the task at hand.
    \item \textbf{Combining program synthesis together with transductive models:} Merging together the two approaches above into a single super-approach, based on the observation that each approach tends to solve different kinds of tasks.
\end{itemize}

The first exciting accomplishment of the year came from Ryan Greenblatt, who reached 42\% on the ARC-AGI-Pub leaderboard~\cite{greenblatt2024arc} using an LLM-guided program synthesis approach. His solution uses GPT-4o to search thousands of Python programs per task (and iteratively debug the most promising ones) to find a program that successfully maps the task input/output examples.

During the 5-month contest period, one of the top scoring teams, MindsAI, improved its score on ARC-AGI-1 private evaluation set from 33\% (achieved by the same team at the end of the 2023 competition) to 55.5\%. MindsAI pioneered test-time training for ARC-AGI, beginning in 2023. While they chose not to publicly share their 2024 TTT implementation, they inspired many teams to invent their own.

Notably, ARC Prize 2024 1st place winners, the ARChitects, used TTT to score 53.5\% on the private evaluation and the 2nd place paper award winners, Ekin Akyürek and team, used TTT to score 47.5\% on semi-private evaluation set. Both were open sourced as a result of the competition and are available on \href{https://arcprize.org}{arcprize.org}.

\pgfplotsset{compat=1.18}

\begin{figure}[h]
    \centering
    \begin{tikzpicture}
        \begin{axis}[
            width=0.8\textwidth,
            height=6cm,
            xlabel={Year},
            ylabel={High Score (\%)},
            ymin=0, ymax=100,
            xtick=      {2019, 2020, 2021, 2022, 2023, 2024, 2025},
            xticklabels={2019, 2020, 2021, 2022, 2023, 2024, 2025},
            ytick={0,20,40,60,80,100},
            legend pos=north west,
            grid=both,
            grid style={line width=.1pt, draw=gray!10},
            major grid style={line width=.2pt,draw=gray!50},
        ]
            \addplot[color=blue, mark=*] coordinates {
                (2019,0)
                (2020.41,20.6)
                (2022.16,28.5)
                (2023.16,30.45)
                (2023.33,31.4)
                (2024.41,34.4)
                (2024.91,55.5)
            };
            \addlegendentry{ARC-AGI-1 Private Eval High Score}
            
            % Manual node placement - you can adjust the y-coordinates to prevent overlap
            \node at (axis cs:2020.41,28) {20.6\%};
            \node at (axis cs:2022.16,38) {28.5\%};
            \node at (axis cs:2023.33,43) {31.4\%};
            \node at (axis cs:2024.41,49) {34.3\%};
            \node at (axis cs:2024.91,65) {55.5\%};
        \end{axis}
    \end{tikzpicture}
    \caption{ARC-AGI-1 high scores over time}
    \label{fig:line_chart}
\end{figure}
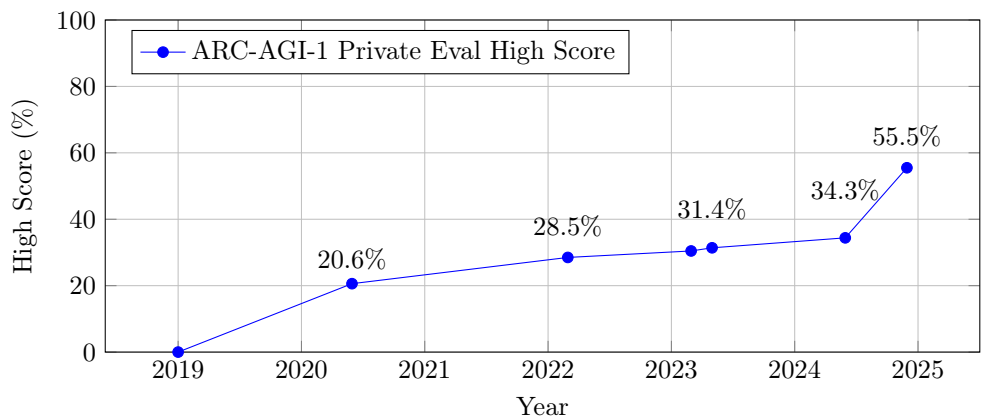

\subsection{Deep Learning-Guided Program Synthesis}

Upon releasing ARC-AGI in 2019, Chollet suggested that it could be understood as a program synthesis benchmark, and that it could be solved by leveraging deep learning models to guide a discrete program search process -- thereby solving the bottleneck of program synthesis, combinatorial explosion. Such an approach was described in detail in Chollet's AAAI Fall Symposium talk in November 2020~\cite{chollet2020aaai}. The 2020 competition was entirely dominated by brute-force program search technique, and the rise of LLMs capable of generating code from 2023 onwards led to more efficient program synthesis solutions that used LLMs to write candidate programs that would then be evaluated by a code interpreter.

Program synthesis for ARC-AGI has so far come in the following flavors:

\begin{itemize}
    \item \textbf{Brute-force search over a Domain Specific Language (DSL)}: This approach involves exhaustively searching the space of possible programs within a predefined DSL. While theoretically complete, it cannot scale on its own to complex programs as it suffers from combinatorial explosion as the size of the DSL and the size of the desired program grow. This is the first approach to have yielded positive results on ARC-AGI, and as early as 2020 we had a proof of existence of a very simple, relatively low-compute brute-force strategy that could achieve 49\% on the private evaluation set by ensembling all 2020 competition entries together. The highest score seen on any single Kaggle submission with this approach has been 40\% on the private evaluation set (by Agnis Liukis, team name alijs.)

    \item \textbf{LLM-powered program generation in open-ended languages}: LLMs pretrained on programming-related data can be used to generate programs in general-purpose languages like Python. Greenblatt~\cite{greenblatt2024arc} demonstrated an approach that prompted GPT-4o with a task description (containing the task's demonstration pairs) to generate thousands of candidate Python programs to solve the task, which were then run by a code interpreter and selected based on their performance on the demonstration pairs. To work well, this approach requires significant prompt engineering efforts and relies on deterministic evaluation of potentially vast numbers of generated programs.

    \item \textbf{LLM-guided discrete program search over a DSL}: This approach combines some of the strengths of both DSL-based discrete program search and LLMs. Ouellette \cite{ouellette2024neurallyguided} demonstrated this strategy using an LLM to guide the search process within a DSL, effectively reducing the search space and improving efficiency.

    \item \textbf{LLM-powered iterative program debugging}: Instead of generating complete programs from scratch, LLMs can be used to iteratively debug and refine programs that are heuristically generated or close to correct. Greenblatt~\cite{greenblatt2024arc} also explored this approach, using LLMs to identify and rectify errors in the most promising incorrect programs (selected via a heuristic criterion), leading to improved performance on ARC-AGI.

\end{itemize}

One approach that has not been tried so far (likely because it is technically challenging) but that we expect to perform well in the future, is the use of specialist deep learning models to guide the branching decisions of a discrete program search process -- similar to what can be seen in the AlphaProof system from Google DeepMind~\cite{alphaproof2024silvermedal}.

We expect that program synthesis, together with closely related test-time search techniques, will be adopted by every frontier AI system over the next 12 to 24 months. This will result in a stronger need for formal efficiency reporting with benchmark scores in the future, since any search-based approach can always score higher if provided with more compute, and it is thus no longer possible to assign a score to an approach, but only to the combination of an approach plus a compute budget. For instance, we estimate that an 85\% ARC-AGI score could be achieved by an approach like Greenblatt's when generating, evaluating, and debugging approximately 100,000,000 programs per task, which would represent a multi-million dollar compute budget to solve 100 tasks.

We also note that deep learning-guided program synthesis does not currently decisively beat DSL-based brute-force program search -- both score in the 40\% range today with comparable compute budgets. We expect that more efficient program search techniques leveraging deep learning should be able to pull away from brute-force search in the future.

\subsection{Test-Time Training}

The classical deep learning paradigm, exemplified by LLMs from the 2022-2023 period, involves first training a model on a large dataset and then performing inference with a frozen version of the model. However, solving ARC-AGI requires going beyond simple fetching and application of memorized patterns -- it necessitates the ability to adapt to the specific task at hand at test time. This has led to the emergence of test-time training (TTT), also known as test-time fine-tuning (TTFT), as a dominant approach in LLM-based solutions for ARC-AGI. Today, all top LLM-based transduction approaches for ARC-AGI leverage TTT, and there does not exist any \textit{static inference}-style transduction solution that scores above 11\%. This stark gap highlights the inability of the classical deep learning paradigm to generalize to novel tasks.

TTT, in this context, involves fine-tuning a pretrained LLM on the demonstration pairs of each task instance seen at test time, effectively creating a different variant of the base model for each task. The model is then prompted to directly predict the output grid (transduction).

Interestingly, TTT can be seen as conceptually similar to program search, albeit at the opposite end of the memorization/recombination spectrum. Both involve recombining pre-existing building blocks to solve a task. Program search typically employs deep recombination of a small set of generic programming primitives. TTT, on the other hand, utilizes a vast number of specialized ``building blocks'' (vector functions found in the weights of the pretrained LLM) and performs shallow recombination through test-time gradient descent.

Key aspects of the application of TTT to ARC-AGI include:

\begin{itemize}
    \item \textbf{Data augmentation and alternative datasets}: Given the limited size of the ARC-AGI-1 dataset, successful TTT relies heavily on increasing the amount of ARC-AGI-like data available for pretraining, either by using alternative datasets like ARC-Heavy/ARC-Potpourri (a combined dataset comprising 400,000 additional ARC-like tasks released by Ellis et al.~\cite{li2024}) and Re-ARC (a programmatic implementation of the ARC-AGI training set allowing infinite sampling of new instances of the 400 training tasks, released by Hodel~\cite{hodel2024rearc}), or by carefully designed data augmentations. The ARChitects~\cite{thearchitects2024}, for example, introduced novel augmentations and a selection criterion based on the stability of generated solutions under these augmentations.

    \item \textbf{Fine-tuning strategies}: Both LoRA fine-tuning~\cite{hu2021lowrank} and full fine-tuning have been explored for adapting LLMs at test time. Fine-tuning is performed on augmented demonstration pairs derived from the specific test instance considered.

    \item \textbf{Specialized 2D-aware architectures}: Effective TTT often leverages specialized transformer architectures tailored for visual reasoning. This includes employing 2D attention mechanisms or 2D position encodings (as seen in the Puget/NVIDIA ``A 2D nGPT Model for ARC Prize'' \cite{puget2024}) to better capture spatial relationships within the input grids.
\end{itemize}

Examples of TTT used in ARC Prize 2024:

\begin{itemize}
    \item OmniARC (Barbadillo~\cite{barbadillo2024}): A Qwen2.5-0.5B-Instruct model pretrained on multiple program induction tasks and further fine-tuned at test time with augmented ARC data. They combine this with a program synthesis approach for a competitive ensemble.

    \item Akyürek et al.~\cite{akyurek2024}: An 8B parameter model using TTT achieved 53\% accuracy on the public evaluation set.

    \item MindsAI~\cite{mindsai2024}: A Salesforce T5 series model pretrained on the public evaluation set and synthetic data, is further fine-tuned at test time on each private task.
    
    \item ARChitects~\cite{thearchitects2024}: A NeMo-Minitron-8B foundation model with extensive data augmentation and a novel selection criterion based on solution stability under augmentations.

\end{itemize}

A variant of TTT involves searching in the latent space of an LLM as in Bonnet and MacFarlane~\cite{bonnet2024}. This approach uses random search and gradient descent to find better program representations within the model's latent space -- a novel take on test-time adaptation that is neither fine-tuning nor discrete search.

Overall, we expect test-time training will be the primary technique for LLM-based AI systems to improve performance on tasks outside of what they were pretrained on. It's harder to incorporate TTT into production systems compared to program synthesis, so we expect it won't be productionized for a couple of years -- but it, or a derivative, should become popular from 2026 onwards.

\subsection{Combining Program Synthesis with Transduction}

There are broadly two ways to approach ARC-AGI:

\begin{itemize}
    \item \textbf{Program synthesis, or ``induction''}: Based on the demonstration pairs of a test task, find a program or function that appears to turn the input grids into their corresponding output grids, then apply the program to the test input grid(s).
    \item \textbf{Transduction}: Based on the demonstration pairs of a test task and an input grid, directly predict the corresponding output, for instance, by prompting an LLM with the task description and the test input.
\end{itemize}

As soon as transduction-based approaches started scoring above zero (in late 2023, pioneered by Jack Cole and Mohamed Osman) researchers noticed that program search and transduction were able to solve significantly distinct sets of tasks. This issue was later investigated in detail by Li et al. in ``Combining Induction and Transduction for Abstract Reasoning''~\cite{li2024}.

Today, all top scores (e.g., Akyürek and Berman on the public leaderboard, and the ARChitects and Barbadillo on the Kaggle leaderboard) use a combination of transduction and induction. The best transduction-only and induction-only single submissions score around 40\%, so only an ensemble of both can compete for the state of the art.

\section{Future}

We have committed to running ARC Prize annually until the ARC-AGI benchmark is defeated and a public reference solution is shared. ARC Prize 2024 was a large-scale experiment that we consider highly successful, and we aspire to grow ARC Prize from its experimental origins into a durable north star for AGI. We are applying lessons learned during ARC Prize 2024 to inform future versions of both the competition and the benchmark.

\subsection{ARC Prize: 2025 and Beyond}

We're excited that ARC Prize catalyzed significant attention towards new ideas for AGI. We had expected ARC Prize to drive academics, independent researchers, and big labs alike to give renewed attention towards ARC-AGI. But we were surprised to the degree that well-funded AI research startups would change their roadmaps to prioritize beating the benchmark. We are now aware of at least seven distinct efforts to solve ARC-AGI by organizations that have greater than \$1M in funding, including Basis AI (\url{basis.ai}), Tufa Labs (\url{tufalabs.ai}), Agemo (\url{agemo.ai}), and Symbolica (\url{symbolica.ai}).

All of these groups do not share the same incentives (e.g., progress prizes would not be sufficient to incentivize sharing for venture-funded startups and large corporate labs), and we plan to redesign the 2025 edition of the competition to account for this. Our goal is to provide the best directional compass towards AGI across the full spectrum of AI research actors, from academic labs to startups to big labs.

\subsection{ARC-AGI-2}

The ARC-AGI-1 private evaluation set has been unchanged since 2019, and it is known to suffer from a number of flaws. First of all, the private evaluation set is limited to only 100 tasks. These 100 tasks have been used by all four ARC-AGI competitions for reporting intermediate leaderboard scores, and, as a result, on the order of 10,000 private evaluation set scores have been reported to participants so far. This presents a significant risk of overfitting, since each score has the potential to extract a tiny but non-zero amount of information about the content of the hidden tasks. The benchmark’s reliability can be improved by increasing the sample size and using two separate datasets: one for intermediate leaderboard scores (a larger semi-private evaluation set) and another for final scoring (a larger private evaluation set). This approach eliminates the risk of overfitting to the private evaluation set.

Further, while 20\% represented the highest score by any single submission in 2020, an analysis of all 2020 submissions found that 49\% of the private evaluation set was solved by at least one team (all of which were using some variation of brute-force program search.) This suggests that a large fraction of the ARC-AGI-1 dataset is overly susceptible to these kinds of brute-force program search techniques and therefore does not carry a useful signal towards AGI. A sufficient fraction of the dataset has proven interesting enough, and that's why ARC-AGI remains unsolved.

Finally, anecdotal evidence suggests that the different evaluation datasets are not drawn from a consistent human difficulty distribution, which makes score comparisons across evaluations challenging.

To address these issues while retaining the familiar task format of ARC-AGI, we are actively working on ARC-AGI-2 and our intention is to launch the new dataset along with the 2025 competition.

\section{Conclusion}

ARC Prize 2024 was a highly successful experiment -- awareness of the benchmark was raised significantly and several new approaches have emerged, bringing the state of the art from 33\% to 55.5\%. However, ARC-AGI remains undefeated -- still by a considerable margin -- especially considering that a score of 49\% was technically achievable with basic brute-force program search as early as 2020. New ideas are still needed to build AGI. The fact that ARC-AGI survived five months of intense scrutiny with an outstanding \$600,000 grand prize and hundreds of thousands of dollars in additional prizes is strong evidence that the solution does not yet exist. We're inspired, proud, and hopeful that ARC-AGI has played an important role in shifting attention towards new research ideas. Our belief is the team that will eventually build AGI is thinking about ARC-AGI today, and we're committed to stewarding this attention as a north star towards AGI.

\section{Appendix}

\subsection{The ARC-AGI Ecosystem}

ARC Prize has inspired a community of researchers and developers who have contributed valuable tools, datasets, and repositories to support both competition participants and the greater AI community.

\begin{itemize}
    \item \textbf{ARC-DSL} - A domain-specific language for working with ARC-AGI tasks: \href{https://github.com/michaelhodel/arc-dsl}{GitHub Repository}
    \item \textbf{ConceptARC} - Benchmark in the ARC-AGI domain that systematically assesses abstraction and generalization abilities on a number of basic spatial and semantic ``concept groups'': \href{https://github.com/victorvikram/ConceptARC?tab=readme-ov-file}{GitHub Repository}
    \item \textbf{RE-ARC} - A repository to procedurally generate examples for the ARC-AGI training tasks: \href{https://github.com/michaelhodel/re-arc}{GitHub Repository}
    \item \textbf{Bootstrapping ARC} - Tools for generating synthetic ARC-AGI problems: \href{https://github.com/xu3kev/BARC}{GitHub Repository}
    \item \textbf{arcsolver} - A Python library for automatically solving ARC challenges using Claude and object-centric modeling: \href{https://github.com/agemoai/arcsolver}{GitHub Repository}
    \item \textbf{ARC Interactive} - An interactive web-based tool for engaging with ARC-AGI tasks: \href{https://neoneye.github.io/arc/}{Web Tool}
    \item \textbf{Arckit} - Python and command-line tools for easily working with the Abstraction \& Reasoning Corpus: \href{https://github.com/mxbi/arckit}{GitHub Repository}
    \item \textbf{The ARC Game} - UI interface for ARC-AGI Tasks: \href{https://volotat.github.io/ARC-Game/}{Web Interface}
    \item \textbf{ARC Gym} - A data generation framework to help research and develop a solution to the problems of compositional generalization and efficient search: \href{https://github.com/SimonOuellette35/ARC_gym}{GitHub Repository}
    \item \textbf{Open Source Kaggle Notebooks} - Hundreds of publicly shared Kaggle submissions for ARC Prize 2024: \href{https://www.kaggle.com/competitions/arc-prize-2024/code}{Kaggle Competition Code}
\end{itemize}

Numerous other resources are listed in the \href{https://arcprize.org/guide#resources}{official ARC Prize Technical Guide}.

\subsection{Acknowledgments}

ARC Prize builds on the legacy of earlier ARC-AGI competitions: the 2020 competition on Kaggle as well as the 2022 and 2023 ``ARCathons'', which were a collaboration between Fran\c{c}ois Chollet and the Davos-based non-profit AI lab, Lab42. We are grateful to the Lab42 team -- Pascal Kaufmann, Rolf Pfister, Oliver Schmid, and Hansueli Jud -- for their expertise and support in facilitating a smooth transition for the ARC-AGI community. Their contributions were significant in facilitating a bigger, bolder competition and advancing initiatives like ARC-AGI-2.

We would also like to recognize the dedication of past ARCathon participants, who not only championed the benchmark but whose prior work brought new members of the community quickly up to speed. In particular, we thank Michael Hodel, Jack Cole, Mohamed Osman, and Simon Ouellette for their ongoing efforts to develop ARC-AGI solutions. Special recognition is due to Simon Strandgaard for his exceptional role as a community ambassador and prolific open-source contributor.

ARC Prize would have not been possible without the support of the Kaggle team. Kaggle plays a critical role in the Artificial and Machine Learning ecosystem. We would like to thank Addison Howard, Walter Reade, Maggie Demkin and Elizabeth Park for the support throughout the 2024 competition.

Finally, we extend our deepest gratitude to all participants in ARC Prize 2024, especially those who shared their work with the community. Your dedication advances the broader field of AI, bringing us closer to realizing the transformative potential of AGI for humanity.

\bibliographystyle{plain}
\bibliography{references}

\end{document}